\documentclass[10pt,journal,twocolumn]{IEEEtran}
\usepackage{amssymb}
\usepackage{amsmath}
\usepackage{graphicx}
\usepackage{subfig}
\usepackage{multirow}
\usepackage{caption}
\usepackage{times}
\usepackage[english]{babel}
\usepackage{mathptmx}
\usepackage{nccmath}
\usepackage{cite}
\usepackage{amssymb,amsthm}
\usepackage{hyperref}
\usepackage{multirow}
\usepackage{algorithm}
\usepackage{algorithmic}
\usepackage{array}
\ifCLASSOPTIONcompsoc
\else
\fi
\ifCLASSINFOpdf
\else
\fi

\floatname{algorithm}{Algorithm}

\begin{document}
\title{Image Denoising Using Low Rank Minimization With Modified Noise Estimation}

\author{Zahid~Hussain~Shamsi, Hyun Sook Oh %~\IEEEmembership{Member,~IEEE,}
        and Dai-Gyoung~Kim*%~\IEEEmembership{Fellow,~OSA,}
        %and~Jane~Doe,~\IEEEmembership{Life~Fellow,~IEEE}% <-this % stops a space
%        Zahid Hussain Shamsi is with University of the Punjab, Lahore, 54590, Pakistan, currently  enrolled in  Ph.D at Hanyang University (Erica Campus), Seoul,
%426-791, Korea
\IEEEcompsocitemizethanks{\IEEEcompsocthanksitem
\IEEEcompsocthanksitem
%Manuscript received xx xx, xxxx; revised xx xx, xxxx.
%This work was supported by the National Research Foundation of Korea (NRF) funded by the Korean (MEST) (NRF-2011-0026245).
\protect
Hyun Sook Oh is with the Department
of Applied Statistics, Gachon University, Seongnam-Si, 461-701, Korea (email: hoh@gachon.ac.kr ).\hfil\break
$^*$Dai-Gyoung Kim and Zahid Hussain Shamsi are with the Department
of Applied Mathematics, Hanyang University (ERICA), Ansan,
426-791, Korea (email: $^{*}$dgkim@hanyang.ac.kr, zshamsi14@hanyang.ac.kr).
}
\protect% <-this % stops a space

%\thanks{Manuscript received xx xx, xxxx; revised xx xx, xxxx.}
}

\IEEEcompsoctitleabstractindextext{%
\begin{abstract}
%The abstract goes here.
Recently, the application of low rank minimization to image denoising has shown remarkable denoising results which are equivalent or better than those of the existing state-of-the-art algorithms. However, due to iterative nature of  low rank optimization, estimation of residual noise is an essential requirement after each iteration. Currently, this noise is estimated by using the filtered noise in the previous iteration without considering the geometric structure of the given image. This estimate may be affected in the presence of moderate and severe levels of noise.  To obtain a more reliable estimate of residual noise, we propose a modified algorithm (GWNNM) which includes the contribution of the  geometric structure of an image to the existing noise estimation. Furthermore, the proposed algorithm exploits the difference of large and small singular values to enhance the edges and textures during the denoising process. Consequently, the proposed modifications achieve significant improvements in  the denoising results of the existing low rank optimization algorithms.
\end{abstract}

\begin{IEEEkeywords}
 Low rank minimization, nuclear norm, singular value decomposition, residual noise estimation, denoising.

\end{IEEEkeywords}}
\maketitle

\IEEEdisplaynotcompsoctitleabstractindextext

\IEEEpeerreviewmaketitle

\section{Introduction}

%\STATE <text>
% \IF{<condition>} \STATE{<text>} \ELSE \STATE{<text>} \ENDIF
% \FOR{<condition>} \STATE{<text>} \ENDFOR
% \FOR{<condition> \TO <condition> } \STATE{<text>} \ENDFOR
% \FORALL{<condition>} \STATE{<text>} \ENDFOR
% \WHILE{<condition>} \STATE{<text>} \ENDWHILE
% \REPEAT \STATE{<text>} \UNTIL{<condition>}
% \LOOP \STATE{<text>} \ENDLOOP
% \REQUIRE <text>
% \ENSURE <text>
% \RETURN <text>
% \PRINT <text>
% \COMMENT{<text>}
% \AND, \OR, \XOR, \NOT, \TO, \TRUE, \FALSE

\IEEEPARstart{P}{atch} based image denoising in conjunction with the notion of non-locality has led to several state-of-the-art algorithms \cite{buades2005non,dabov2007image,
elad2006image,mairal2009non,Dongetal_SAIST_6319405}.
These algorithms exploit certain prior knowledge like image redundancy and sparse representation of images in suitable transform domains. These assumptions are found to be generally true in case of natural images and greatly improve the denoising results.
%For instance NL-means filtering \cite{buades2005non} exploits the redundant non-local similar patches across the image to estimate image intensity at each pixel location. Furthermore, it has been observed that natural images admit sparse decomposition in certain suitable transform domain depending upon the geometric features of the given images. The sparse representation then  may lead to enhanced results in various computer vision applications, particularly, in case of image denoising. BM3D \cite{dabov2007image}, for example, performs collaborative filtering on 3D groups of similar non-local patches using pre-designed dictionary like Haar or DCT transform to achieve enhanced sparse representation and in turn outstanding denoising results.  Another state-of-the-art algorithm LSSC \cite{mairal2009non} extends an elegant approach of dictionary learning for sparse representation \cite{elad2006image} instead of using pre-designed dictionary. LSSC  exploits  $L_{1,2}$ grouped sparsity regularization for simultaneous sparse coding of similar non-local patches during learning phase and subsequently uses $L_0$ norm for final reconstruction of the denoised image.
\par In addition to sparsity and redundancy priors, a low rank assumption has also been proposed recently in \cite{Dongetal_SAIST_6319405,GuWNNM6909762}. According to low rank prior, a matrix constructed by stacking non-local similar patches from a given noisy image resides in a low dimensional subspace of a high dimensional space and therefore satisfies the low rank criterion. Dong \textit{et al}. \cite{Dongetal_SAIST_6319405} introduced an iterative soft thresholding scheme (SAIST) using $L_{1,2}$ grouped sparsity regularization to penalize the singular values of the low rank matrices. More recently, Gu \textit{et al}. \cite{GuWNNM6909762} have further explored the low rank prior using weighted nuclear norm minimization (WNNM). This algorithm exploits the physical significance of the singular values and treats them differently according to their respective magnitudes. Both low rank minimization algorithms \cite{Dongetal_SAIST_6319405,GuWNNM6909762} have produced outstanding denoising results by exploiting low rank prior confirming to its validity as a suitable assumption for image restoration problem.
\par In this paper, we investigate WNNM algorithm for its  noise estimation scheme. The motivation to consider the noise estimation relies on the fact that the residual noise is one of the key factors for subsequent computation of singular values and corresponding threshold weights which are the indispensable ingredients of low rank minimization schemes,in particular, for WNNM algorithm. WNNM algorithm estimates residual noise by comparing the grouped patch matrices in a given noisy image with the corresponding patch matrices in its denoised version obtained in the previous iteration.
In practice, this difference is intuitively assumed to be the noise. However, this assumption may not be generally true, in particular, for the images with the complex geometric structures \cite{Liu_GNOISE_EST_6466947}.
\par In order to obtain a more reliable estimate of residual noise, we note that the geometric structure of the image should also be taken into account. To the best of our knowledge, this aspect of residual noise estimation for WNNM algorithm has remained unnoticed up to now. By considering the proposed contribution to the existing residual noise estimation, we have obtained remarkable improvements in the denoising results of the original WNNM algorithm for most of the test images. In particular, for moderate and high levels of noise these improvements are quite significant.
\par We also propose to reinforce edges and texture during iterative denoising by splitting the singular value decomposition (SVD) into low and high components. Subsequently, two denoised images are obtained using low and high components of the SVD, respectively. The difference of these two images is used as feedback to reinforce edges and textural regions during iterative denoising process. This modification further enhances the denoising capability of the proposed algorithm.
\par The rest of the paper is organized as follows. We briefly describe low rank minimization algorithm for image denoising in Sec.~\ref{sec:WNNM_2}.  The proposed algorithm and its implementation  are presented in Sec.~\ref{sec:PoposedAlgo_3}. Experimental settings and results are described in Sec.~\ref{sec:results4}. Finally, conclusions are drawn in Sec.~\ref{sec:Conc5}.
\section{Low Rank Minimization for image denoising}\label{sec:WNNM_2}
We consider the well known image denoising problem
\begin{equation}\label{eq:noisyimage}
{\bf y}={\bf x}+{\bf n},
\end{equation}
where ${\bf x}$ is the original image to be recovered from the noisy observed image ${\bf y}$ and ${\bf n}\sim \mathcal{N}\left(0,\sigma_n\mathbb{I}\right)$ is an additive white Gaussian noise with known standard deviation $\sigma_n$.  In terms of patch based formulation, non-local similar patches are searched for local patch ${\bf p_{y_j}}$ at each pixel location $j$ in the observed image.  Afterwards, the patches similar to a patch ${\bf p_{y_j}}$  can be stacked to construct a matrix $\bf M_{y_j}$. Using patch based formulation, the original problem (\ref{eq:noisyimage}) can be expressed as \cite{GuWNNM6909762}
\begin{equation}\label{eq:MatrixFormofProblem}
\bf M_{y_j}=\bf M_{x_j}+\bf M_{n_j},
\end{equation}
where ${\bf M_{x_j}}$ and ${\bf M_{n_j}}$ denote the corresponding patch matrices of original image and additive noise, respectively.
\par The patch matrix ${\bf M_{x_j}}$ defined in (\ref{eq:MatrixFormofProblem}) is intuitively assumed to be a low rank matrix because of the image redundancy and similarity prior \cite{Dongetal_SAIST_6319405}. Thus, a noise free patch can be recovered by using low rank minimization. However,
low rank minimization is a non-convex NP hard problem \cite{CaiCandes_SVDThr}
\cite{Candes:2011:RPC:1970392.1970395}. Therefore, nuclear norm minimization
(NNM) has been heuristically proposed as its convex regularization which can be written as \cite{GuWNNM6909762,CaiCandes_SVDThr}
\begin{equation}\label{eq:NNM_form}
{\bf \hat{M}_{x_j}}=\arg \min_{\bf {M}_{x_j}}
\frac{1}{\sigma_n^2} \left\|\bf {{M}_{y_j}}-\bf{{M}_{x_j}}\right\|_F^2
+\left\|\bf{{M}_{x_j}}\right\|_{*},
\end{equation}
where $\left\|.\right\|_{*}=\sum_i\left|\lambda_i\left(.\right)\right|$ is the nuclear norm defined using the singular values $\lambda_i$ of the matrix $\bf{{M}_{x_j}}$ and $\left\|.\right\|_F$ denotes Frobenius norm.
Convexity of the reformulated problem (\ref{eq:NNM_form}) assures the global optimal solution through singular value decomposition  \cite{CaiCandes_SVDThr}
\begin{eqnarray}\label{eq:NNM_solution}
\nonumber U\Lambda V^T&=&svd\left({\bf \hat{M}_{x_j}}\right),\\
\Lambda_\theta&=&S_\theta\left(\Lambda\right)=\max(\lambda_i-\theta,0),
\end{eqnarray}
where $\theta$ denotes a soft threshold applied to all the singular values $\lambda_i$ of the matrix $\bf{{M}_{x_j}}$ without considering the importance of the large and small singular values.
\par To consider physical significance of singular values, Gu \textit{et al}. \cite{GuWNNM6909762} proposed weighted nuclear norm minimization (WNNM) algorithm which replaces $\left\|.\right\|_{*}$ by $\left\|.\right\|_{w,*}$ in (\ref{eq:NNM_form})
%begin{equation}\label{eq:WNNM_form}
%{\bf \hat{M}_{x_j}}=\arg \min_{\bf {M}_{x_j}}
%\frac{1}{\sigma_n^2} \left\|\bf {{M}_{y_j}}-\bf{{M}_{x_j}}\right\|_F^2
%+\left\|\bf{{M}_{x_j}}\right\|_{w,*},
%\end{equation}
where $\left\|.\right\|_{w,*}=\sum_i\left|w_i\lambda_i\left(.\right)\right|$ is the weighted nuclear norm and the weights $w_i$ are defined as \cite{GuWNNM6909762}
\begin{equation}\label{eq:weightedthreshold}
w_i=\frac{c\sqrt{m}}{\left(\lambda_i+\epsilon\right)},
\end{equation}
where $c>0$ is a constant and $m$ is the number of non-local patches similar to the given patch ${\bf p_{y_j}}$. To avoid possible division by zero, we set the constant $\epsilon=10^{-16}$. Note that in this case, the solution is similar to (\ref{eq:NNM_solution}) with the exception that $\Lambda_\theta$ is now written as \begin{equation}\label{eq:WNNM_solution}
\Lambda_{\bf w}\equiv S_{\bf w}\left(\Lambda\right)=\max\left(\lambda_i-w_i,0\right).
\end{equation}
It can be observed from (\ref{eq:weightedthreshold}) and (\ref{eq:WNNM_solution}), the larger singular values are less penalized as compared to the smaller ones in accordance with the basic requirement of the image denoising. However, the global optimal solution is not guaranteed because WNNM becomes a non-convex problem due to the non-descending threshold weights $w_i$ \cite{GuWNNM6909762}. Nevertheless, Gu \textit{et al}. \cite{GuWNNM6909762} have shown that iteratively, this optimization scheme reaches its local minimum fixed point solution.
\section{Proposed Algorithm}\label{sec:PoposedAlgo_3}
In the following, we describe how the existing residual noise estimation used in \cite{Dongetal_SAIST_6319405,GuWNNM6909762} can be further enhanced by considering the proposed modification.
\subsection{Proposed Noise Estimation Method}
Due to the pivotal role of residual noise, it is highly desirable to quantify the amount of noise left in the current denoised image for further processing in the next iteration. This remaining noise is termed as residual noise in this paper.
%Furthermore, the computation of the singular values $\lambda_i$'s and subsequently, the computation of threshold weights $w_i$'s depend upon the standard deviation of the residual noise. Therefore a reliable estimation of residual noise is a critical requirement at each iterative step.
The variance of the residual noise for next iteration $(k+1)$ can be defined as the difference between the variance of the initial given white Gaussian noise $\sigma_n^2$ and that of the filtered noise $\left(\sigma_{flt}^{(k)}\right)^2$ at previous iteration $k$. In WNNM algorithm, the variance of the filtered noise is estimated by comparing grouped patches matrix at location $j$ in the given noisy image ${\bf y}$ and the corresponding matrix in its previously denoised version ${\bf y}^{(k)}$. For simplicity, the variance of filtered noised can be expressed at image level as \cite{Dongetal_SAIST_6319405}
\begin{equation}\label{eq:varianceFilteredNoise}
\left(\sigma_{flt}^{(k)}\right)^2= \left\|{\bf y}-{\bf y}^{(k)}\right\|_{l_2}.
\end{equation}
By using (\ref{eq:varianceFilteredNoise}), Dong \textit{et al}.  \cite{Dongetal_SAIST_6319405} proposed an intuitive estimate for the standard deviation of the residual noise as
\begin{equation}\label{eq:residualNoise_WNNM}
\sigma_{res}^{(k+1)}=\gamma\sqrt{\sigma_n^2-
\left(\sigma_{flt}^{(k)}\right)^2},
\end{equation}
where $\sigma_{res}^{(k+1)}$ denotes the estimated standard deviation of the residual noise present at $(k+1)^{th}$ iteration. The constant $\gamma>0$ is a scaling factor, heuristically introduced to control the re-estimation of the standard deviation \cite{Dongetal_SAIST_6319405}. Using this noise estimate, the singular values of local patch matrix ${\bf M_{y_j}}^{(k+1)}$ are adjusted as \cite{GuWNNM6909762}
\begin{equation}\label{eq:EstimatedSingularValues}
\lambda_i\left(\!{\bf M_{y_j}}^{\!(k+1)}\!\right)=
\sqrt{\max\left(\lambda_i^2\left({\bf M_{y_j}}^{\!(k)}\right)\!-m\left(\sigma_{res}^{\!(k+1)}\right)^2,0\right)}.
\end{equation}
It is important to note that the noise estimation approach (\ref{eq:residualNoise_WNNM}) utilizes $l_2$ norm (\ref{eq:varianceFilteredNoise}) which is sensitive to outliers, more specifically, in the presence of severe noise. More importantly, the difference (\ref{eq:varianceFilteredNoise}) between a given noisy image and its filtered version is intuitively taken as noise. However, this assumption may not be generally true, in particular, for the images with the complex geometric structures \cite{Liu_GNOISE_EST_6466947}. In fact, the filtered noise also contains some geometric structure which is lost during denoising. Thus, the above expression of residual noise (\ref{eq:residualNoise_WNNM}) needs certain modification to account for the geometric details of the previously denoised image for residual noise estimation. In order to identify the image structure in terms of patches, Zhu and Milanfar \cite{ZHU_MILANFAR_WEAKTEXTURE5484579} proposed the diagonalization of  gradient covariance matrix $C$ for each patch ${\bf p_{y_j}}$. The covariance matrix $C$ is defined as \cite{ZHU_MILANFAR_WEAKTEXTURE5484579}
\begin{equation}\label{eq:GradientCovarianceMat}
C=G^T G=\sum_{l\in {\bf p_{y_j}}}\left(\begin{array}{cc}
 g_x^2(l) & g_x(l)g_y(l) \\
 g_y(l)g_x(l) &  g_y^2(l)
\end{array}\right),
\end{equation}
where $[g_x(l),g_y(l)]^T$ denotes the gradient vector at location $l$ within the patch ${\bf p_{y_j}}$. The large and small eigenvalues of the covariance matrix $C$ represent the directions of maximum and minimum variations in the image structure, respectively.
Yan-Li \textit{et al}.  \cite{Liu_GNOISE_EST_6466947} utilized this concept of gradient covariance matrices to identify the weak textured patches for the estimation of the noise variance in a given noisy image. The standard deviation, $\sigma_{geom}^{(k)}$, of the noise is subsequently estimated by applying PCA to these weak textured patches iteratively. We propose to consider this geometric contribution in residual noise estimation as
\begin{equation}\label{eq:residualNoise_global}
\sigma_{res}^{'(k+1)}=\gamma\sqrt{\sigma_n^2-
\left(\sigma_{geom}^{(k)}\right)^2}.
\end{equation}
Finally, the modified noise estimate is expressed as a convex combination of geometric (\ref{eq:residualNoise_global}) and non-geometric contributions (\ref{eq:residualNoise_WNNM})
\begin{equation}\label{eq:combinedNoiseEstimate}
\hat{\sigma}_{res}^{(k+1)}= \alpha\sigma_{res}^{(k+1)}+(1-\alpha)\sigma_{res}^{'(k+1)},
\end{equation}
\begin{table*}[ht!b]
\centering
\caption{Denoisig results (PSNR) of various state-of-the-art methods
}
\renewcommand{\arraystretch}{1.3}
{\scriptsize \begin{tabular}{!{\vrule width 1pt}c!{\vrule width 1pt}c|c|c|c|c!{\vrule width 1pt}c|c|c|c|c!{\vrule width 1pt}c|}\noalign{\hrule height 1pt}
&\multicolumn{5}{c|}{${\bf \sigma_n=10}$}& \multicolumn{5}{|c!{\vrule width 1pt}}{${\bf\sigma_n=30}$}\\ \noalign{\hrule height 1pt}
{\bf Image}&{\bf BM3D}&{\bf LSSC}&{\bf SAIST}&
{\bf WNNM}&{\bf GWNNM}&{\bf BM3D}&{\bf LSSC}&{\bf SAIST}&{\bf WNNM}&{\bf GWNNM}\\\noalign{\hrule height 1pt}
C. Man&34.18&34.24&34.30&34.44&\textbf{34.45}&28.64&28.63&28.36&\textbf{28.80}&\textbf{28.80}\\\hline
House&36.71&36.95&36.66&36.95&\textbf{36.98}&32.09&32.41&32.30&35.52&\textbf{35.57}\\\hline
Peppers&34.68&34.80&34.82&34.95&\textbf{34.97}&29.28&29.25&29.24&29.49&\textbf{29.51}\\\hline
Monarch&34.12&34.44&34.76&\textbf{35.03}&\textbf{35.03}&28.36&28.20&28.65&28.92&\textbf{28.93}\\\hline
J.Bean&37.91&38.69&38.37&38.93&\textbf{38.97}&31.97&32.39&32.14&32.46&\textbf{32.56}\\\hline
Lena&35.93&35.83&35.90&36.03&\textbf{36.04}&31.26&31.18&31.27&31.43&\textbf{31.50}\\\hline
Barbara&34.98&34.98&35.24&\textbf{35.51}&35.49&29.81&29.60&30.14&30.31&\textbf{30.43}\\\hline
F.print&32.46&32.57&32.69&\textbf{32.82}&\textbf{32.82}&26.83&26.68&26.95&26.99&\textbf{27.10}\\\hline
Boat&33.92&34.01&33.91&\textbf{34.09}&34.08&29.12&29.06&28.98&\textbf{29.24}&29.21\\\hline
Hill&33.62&33.66&33.65&\textbf{33.79}&\textbf{33.79}&29.16&29.09&29.06&\textbf{29.25}&29.22\\\hline
Man&33.98&34.10&34.12&34.23&\textbf{34.25}&28.86&28.87&28.81&\textbf{29.00}&28.99\\
%\hline Couple&34.04&34.01&33.96&\textbf{31.14}&\textbf{34.14}&28.87&28.77&28.72&\textbf{28.98}&28.96\\
\noalign{\hrule height 1pt}
&\multicolumn{5}{c|}{${\bf \sigma_n=50}$}& \multicolumn{5}{|c!{\vrule width 1pt}}{${\bf\sigma_n=100}$}\\ \noalign{\hrule height 1pt}
{\bf Image}&{\bf BM3D}&{\bf LSSC}&{\bf SAIST}&
{\bf WNNM}&{\bf GWNNM}&{\bf BM3D}&{\bf LSSC}&{\bf SAIST}&{\bf WNNM}&{\bf GWNNM}\\\noalign{\hrule height 1pt}
C. Man&26.12&26.35&26.15&\textbf{26.42}&\textbf{26.43}&23.07&23.15&23.09&23.36&\textbf{23.39}\\\hline
House&29.69&29.99&30.17&30.32&\textbf{30.42}&25.87&25.71&26.53&26.68&\textbf{26.82}\\\hline
Peppers&26.68&26.79&26.73&26.91&\textbf{26.93}&23.39&23.20&23.32&23.46&\textbf{23.56}\\\hline
Monarch&25.82&25.88&26.10&\textbf{26.32}&26.31&22.52&22.24&22.61&22.95&\textbf{22.99}\\\hline
J.Bean&29.26&29.42&29.32&29.62&	 \textbf{29.71}&25.80&25.64&25.82&26.04&\textbf{26.15}\\\hline
Lena&29.05&	 28.95&29.01&29.24&\textbf{29.34}&25.95&25.96&25.93&26.20&\textbf{26.34}\\\hline
Barbara&27.23&27.03&27.51&27.79&\textbf{27.93}&23.62&23.54&24.07&24.37&\textbf{24.49}\\\hline
F.print&24.53&24.26&24.52&24.67&\textbf{24.76}&21.61&21.30&21.62&21.81&\textbf{21.87}\\\hline
Boat&26.78&	26.77&26.63&\textbf{26.97}&26.93&23.97&23.87&23.80&24.10& \textbf{24.15}\\\hline
Hill&27.19&27.14&27.04&\textbf{27.34}&27.31&24.58&24.47&24.29&\textbf{24.75}&24.71\\\hline
Man&26.81&26.72&26.68&\textbf{26.94}&26.92&24.22&23.98&24.01&\textbf{24.36}&\textbf{24.36}\\
%\hline Couple&26.46&26.35&26.30&\textbf{26.65}&	 26.61&23.51&23.27&23.21&\textbf{23.55}&23.53\\
\noalign{\hrule height 1pt}
 \end{tabular}}\label{tb:ComparisonPSNR}
 \end{table*}
where $\alpha\in(0,1)$  is tuned experimentally according to the initially given standard deviation $\hat{\sigma}_{res}^{(0)}=\sigma_n$. For low levels of initial noise $(\sigma_n<30)$, we choose $\alpha=0.9$ and for moderate and severe noise levels $(\sigma_n\ge30)$, we set $\alpha=0.8$.
%*********************************************
\begin{figure}[b]
\centering
%\subfloat[][] \includegraphics[width=1.5in,height=1.5in]{nvcomparison}
%\subfloat[][Intensity based patches]{\includegraphics[width=1.0in,height=1.0in]{figure1.pdf}}
\subfloat[][] {\includegraphics[width=0.8in,height=0.8in]{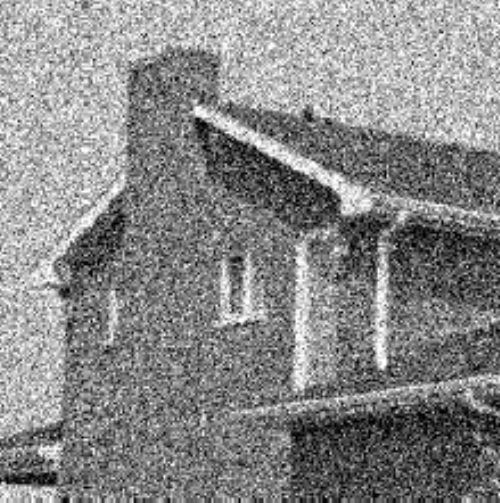}}
\label{fig:f1_house_sn50}~
\subfloat[][] {\includegraphics[width=0.8in,height=0.8in]{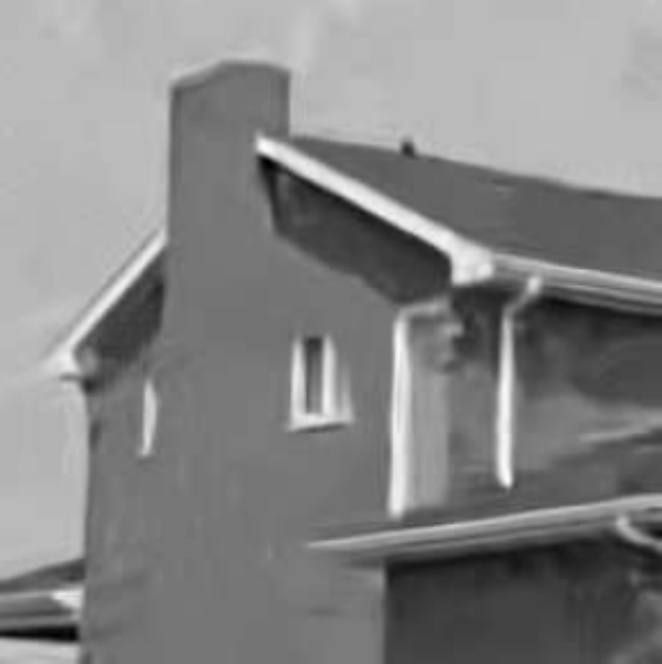}}
\label{fig:f2_houseBm3d_sn50}~
\subfloat[][] {\includegraphics[width=0.8in,height=0.8in]{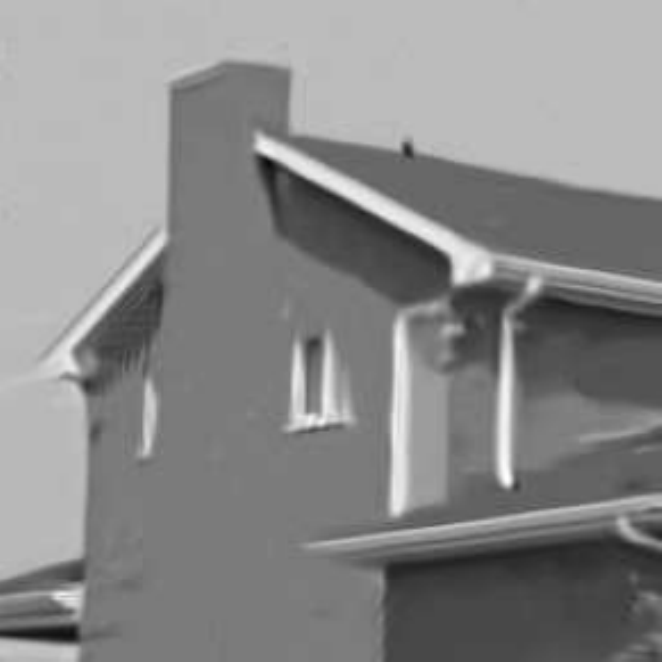}}
\label{fig:f3_houseWNNM_sn50}~
%\subfloat[][] {\includegraphics[width=0.8in,height=0.8in]{figure4.pdf}}
%\label{fig:f4_houseGWNNM_sn50}\\
\subfloat[][] {\includegraphics[width=0.8in,height=0.8in]{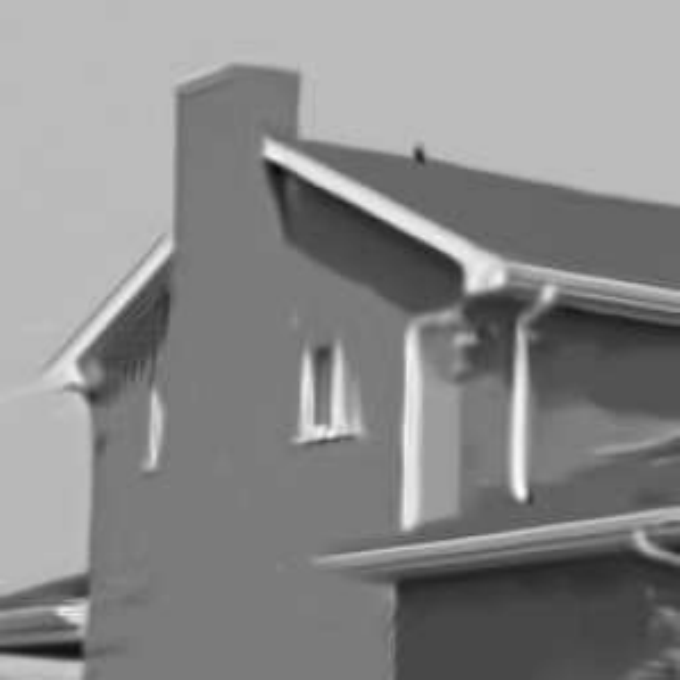}}
\label{fig:f4_houseGWNNM_sn50}\\
\subfloat[][] {\includegraphics[width=0.8in,height=0.8in]{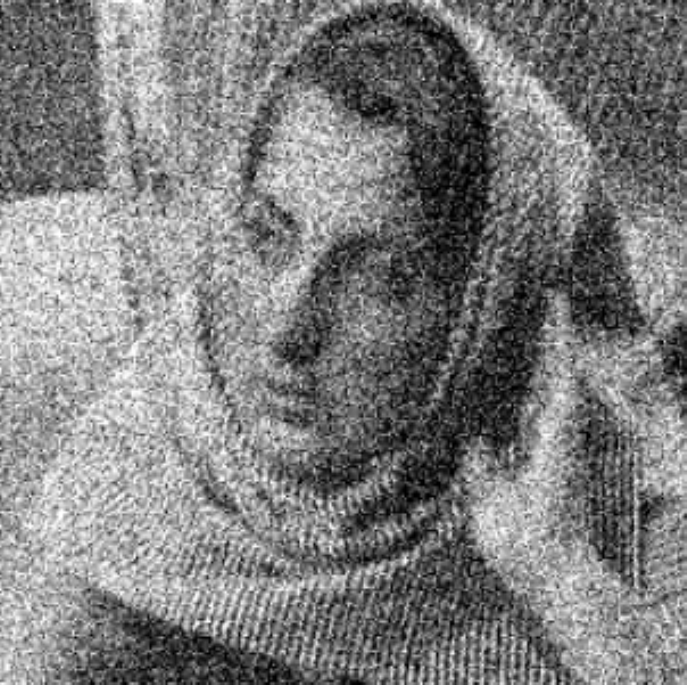}}
\label{fig:f5_barbara_sn50}~
\subfloat[][] {\includegraphics[width=0.8in,height=0.8in]{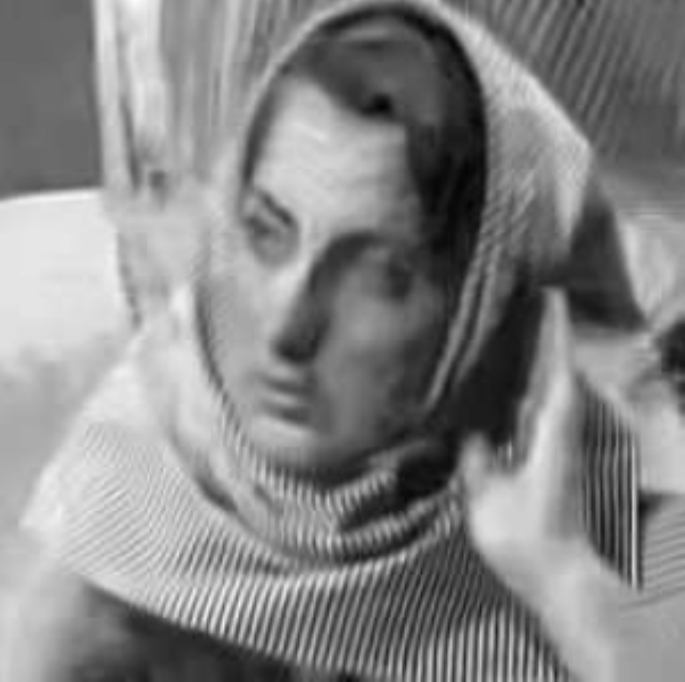}}
\label{fig:f6_barbaraBm3d_sn50}~
\subfloat[][] {\includegraphics[width=0.8in,height=0.8in]{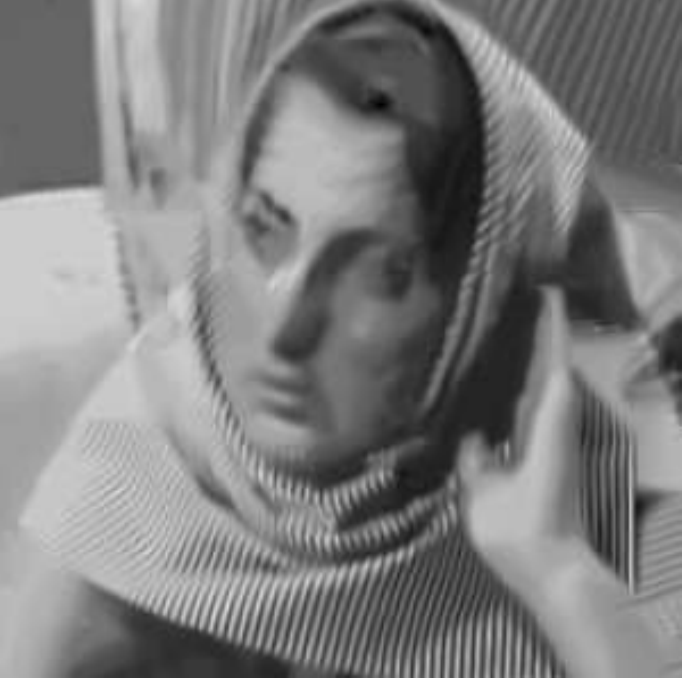}}
\label{fig:f7_barbaraWNNM_sn50}~
\subfloat[][] {\includegraphics[width=0.8in,height=0.8in]{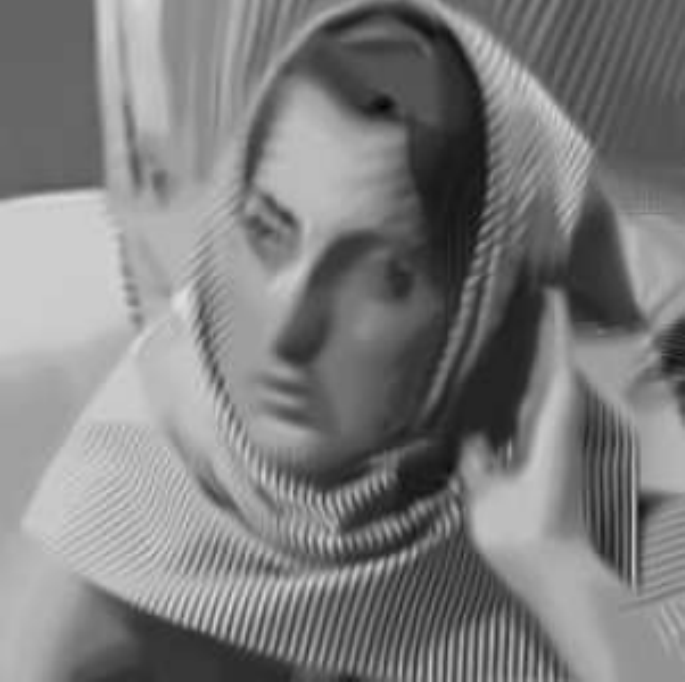}}
\label{fig:f8_barbaraGWNNM_sn50}
\caption{Columns from left to right show the noisy ($\sigma_n=50$) and denoised images using BM3D, WNNM and GWNNM for house and Barbara images, respectively. }\label{fig:house_barbara_n50}
\end{figure}
%*********************************************
The relatively smaller value of $\alpha$  for $\sigma_n\ge30$ indicates that the previous noise estimation (\ref{eq:residualNoise_WNNM}) is significantly affected due to higher levels of  given noise and is subjected to larger correction from the geometric counterpart. Fine tuning of $\alpha$ is also possible according to different noise levels. However, our primary objective is to highlight the importance of the proposed modification. Subsequently, the singular values can now be adjusted by plugging (\ref{eq:combinedNoiseEstimate}) into (\ref{eq:EstimatedSingularValues}).
%\begin{equation}\label{eq:EstimatedSingularValuesNewOne}
%\lambda_i\left({\bf M_{y_j}}^{(k+1)}\right)=
%\sqrt{\max\left(\lambda_i^2\left({\bf M_{y_j}}^{(k)}\right)-m\left(\sigma_{n}^{(k+1)}\right)^2,0\right)}.
%\end{equation}
\subsection{Edge and Texture Enhancement}
Due to thresholding (\ref{eq:WNNM_solution}), some of the smaller singular values may eventually reduce to zero or very close to zero. However, we propose to utilize these values in the feedback step for edge and texture reinforcement as follows. We identify large and small singular values by setting a threshold $\tau$ and then split the truncated matrix $\Lambda_{\bf w}$ as
\begin{equation}\label{eq:splitLambda}
\Lambda_{\bf w}=\Lambda_{\bf w}^{high}+\Lambda_{\bf w}^{low},
\end{equation}
where $\Lambda_{\bf w}^{(high)}$ and $\Lambda_{\bf w}^{(low)}$  contain only large and small singular values of ${\bf M_{y_j}}$, respectively. The optimal solution (\ref{eq:WNNM_solution}) can then be re-written as
\begin{eqnarray}
{\bf M_{y_j}}&=& U\left(\Lambda_w^{(high)}+\Lambda_w^{(low)}\right)V^T
={\bf M_{y_j}}^{(high)}+{\bf M_{y_j}}^{(low)}.\label{eq:splitPatchMatrixM}
\end{eqnarray}
As a result of this splitting, we construct two corresponding images ${\bf {y}}^{(high)}$ and ${\bf {y}}^{(low)}$ which contain high and low energy components, respectively. By exploiting intrinsic nature of these two images, the difference image ${\bf {y}}^{(high)}- {\bf {y}}^{(low)}$  can be associated to sharpness and contrast enhancement. Also the noise level is gradually reduced in these two images, iteratively. Thus, the edges and texture may become more clear in the difference image at each iteration. Therefore, we intuitively propose to use this difference in the feedback step to preserve and enhance geometric details of the image during iterative denoising. Experimental results confirm the effectiveness of the proposed modification as well.
\par The summary of the proposed algorithm is provided in Algorithm \ref{alg:algo1}.
\begin{algorithm}
\caption{Image Denoising Using GWNNM}
     \begin{algorithmic}[1]\label{alg:algo1}
     \REQUIRE {Noisy image ${\bf y}$}
        \STATE{Initialize %$\hat{\bf x}^{(0)}={\bf y}$,
        ${\bf y}^{(1)}={\bf y}$,
        ${\bf y}_{low}^{(1)}={\bf y}$,
        ${\bf y}_{high}^{(1)}={\bf y}$}
            \FOR{k=1:L-1}
                \STATE{$
                {\bf y}^{(k+1)}=
                {\bf y}^{(k)}
                +\delta\left({\bf y}-{\bf y}^{(k)}\right)
                +\eta\left({\bf y}_{high}^{(k)}-{\bf y}_{low}^{(k)}\right)
                $}
                \FOR{each patch ${\bf p_{y_j}}\in {\bf y}^{(k+1)}$}
                \STATE{Construct patch matrix ${\bf M_{y_j}}$ using
                (\ref{eq:MatrixFormofProblem})}
                \STATE{Estimate residual noise $\hat{\sigma}_{res}^{(k+1)}$ using (\ref{eq:combinedNoiseEstimate})}
                \STATE{Compute $\left[ U, \Lambda, V\right]=
                SVD\left({\bf M_{y_j}}\right)$}
                \STATE{Compute threshold weight vector ${\bf w}$ using (\ref{eq:weightedthreshold})}
                \STATE{Obtain truncated matrix $\Lambda_{\bf w}$ using (\ref{eq:WNNM_solution})}
                \STATE{Split
                $\Lambda_{\bf w}=\Lambda_{\bf w}^{low}+\Lambda_{\bf w}^{high}$
                using threshold $\tau$ and (\ref{eq:splitLambda})}
                \STATE{Obtain the estimates ${\bf M_{y_j}}^{low}$ and ${\bf M_{y_j}}^{high}$ using (\ref{eq:splitPatchMatrixM})}
                \ENDFOR
            \STATE{Aggregate ${\bf M_{y_j}}$, ${\bf M_{y_j}}^{low}$ and
            ${\bf M_{y_j}}^{high}$ to construct images ${\bf y}^{(k+1)}$, ${\bf y}_{low}^{(k+1)}$ and ${\bf y}_{high}^{(k+1)}$, respectively}
           % \STATE{Aggregate ${\bf M_{y_j}}^{low}$ to construct the image ${\bf y}_{low}^{(k+1)}$}
%            \STATE{Aggregate ${\bf M_{y_j}}^{high}$ to construct the image  ${\bf y}_{high}^{(k+1)}$}
            \ENDFOR
            \ENSURE{Clean Image ${\bf y}^{(L)}$}
     \end{algorithmic}
\end{algorithm}
\section{Experimental Results}\label{sec:results4}
We consider eleven frequently used standard images to compare the proposed algorithm (GWNNM) with various state-of-the-art algorithms: BM3D \cite{dabov2007image}, LSSC \cite{mairal2009non}, SAIST \cite{Dongetal_SAIST_6319405}, WNNM \cite{GuWNNM6909762}. We use PSNR values for comparison which are reported in \cite{GuWNNM6909762}.
\par We use the same default values of the parameters which were selected in WNNM, heuristically, depending upon the initially given noise level \cite{GuWNNM6909762}. Therefore we skip the details of those parameters due to space limitation. However, we have set the size of search window ($\Delta_w=15\times15$) instead of ($30\times30$) for all the images of size $512\times512$.  In fact, block matching process in WNNM is computationally very expensive and in case of $\Delta_w=30\times30$  for $512\times512$ images, the computational cost is much greater than that of $256\times256$ image. Furthermore, it is pointed out in \cite{salmon2010two} that selecting larger size of search window has no major benefit in denoising results except for pseudo-periodic images like fingerprint image. Therefore, for fingerprint image we use the same value of $\Delta_w=30\times30$ as used in WNNM. The setting of new parameter $\alpha$ has already been explained in the previous section. The other two parameters $\eta$ and $\tau$ are experimentally selected as  $0.01$ and $0.5$, respectively.
\par The comparison of the denoising results is described in  Table.~\ref{tb:ComparisonPSNR} and the best results are highlighted using bold faced values.  Also, a limited comparison of visual quality for BM3D, WNNM and GWNNM algorithms is  shown in Fig.~\ref{fig:house_barbara_n50} at noise level $\sigma_n=50$. As shown in Table.~\ref{tb:ComparisonPSNR} that WNNM and GWNNM perform always better than the rest of the algorithms. For low noise ($\sigma_n=10$) the performance of GWNNM is in general equivalent to WNNM. It is reasonable because the filtered noise (\ref{eq:varianceFilteredNoise}) contains negligible signatures of geometric details and needs only a small correction from geometric estimate (\ref{eq:residualNoise_global}).
%the old residual noise estimate (\ref{eq:residualNoise_WNNM}) needs only a small correction at low noise level.
However, for high noise levels ($\sigma_n\ge30$), GWNNM generally performs  better than WNNM. In particular, at $\sigma_n=100$, the improvements in PSNR values are greater than $0.1$ dB for the images of house, jelly beans, peppers, Lena, Barbara and fingerprint.  Thus, the improved results, in particular, for geometrically rich and complex images like Barbara, fingerprint, house and Lena, support our assumption that the estimation of residual noise needs modification due to geometric structure of the image.
\section{Conclusion}\label{sec:Conc5}
We have highlighted the importance of residual noise estimation for low rank optimization to further improve the results. Further we have proposed a method to make the noise estimate more precise by using geometrical structure of a given noisy image. Alternate effective approaches for noise estimation also exist and can be utilized. However, our primary intent was to indicate the deficiency in the existing residual noise estimation scheme which only exploits filtered noise in the previous iteration without considering the geometric details present in the filtered noise.
%\acknowledgments

\bibliographystyle{ieeetr}
\bibliography{LRM_Denoising}

\end{document}